\documentclass[letterpaper, 10 pt, conference]{ieeeconf} 

\IEEEoverridecommandlockouts                        

\usepackage{graphics}
\usepackage{amsmath} 
\usepackage{amssymb}
\usepackage{cite}
\usepackage{tikz}
\usepackage{tabularx}
\usepackage{array}
\usepackage{booktabs}
\usepackage{bm}
\usepackage{algorithm}
\usepackage{algorithmic}
\usepackage{bbm}
\usepackage{adjustbox}
\usepackage{nicematrix}
\usepackage{caption}
\usepackage{fontawesome5}
\usepackage{tablefootnote}
\usepackage{hyperref}
\usepackage{flushend}

\definecolor{green}{RGB}{11,155,13}

\newcommand{\wmvct}{\textsc{wmvct}}
\newcommand{\tal}{\textsc{tal}}
\newcommand{\coder}{\textsc{VertiCoder}}

\DeclareRobustCommand{\mask}{%
  \begingroup\normalfont
  \includegraphics[height=\fontcharht\font`\B]{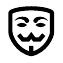}%
  \endgroup
}

\title{\LARGE \bf
VertiCoder: Self-Supervised Kinodynamic Representation Learning on Vertically Challenging Terrain
}

\author{Mohammad Nazeri$^1$, Aniket Datar$^1$, Anuj Pokhrel$^1$, Chenhui Pan$^1$, Garrett Warnell$^{2,3}$, and Xuesu Xiao$^1$
\thanks{$^1$Department of Computer Science, George Mason University {\tt\scriptsize \{mnazerir, adatar, apokhre, cpan7, xiao\}@gmu.edu} $^2$DEVCOM Army Research Laboratory {\tt\scriptsize garrett.a.warnell.civ@army.mil} $^3$Department of Computer Science, The University of Texas at Austin}
}

\begin{document}
\maketitle
\thispagestyle{empty}
\pagestyle{empty}

\begin{abstract}
We present \coder, a self-supervised representation learning approach for robot mobility on vertically challenging terrain. Using the same pre-training process, \coder~can handle four different downstream tasks, including forward kinodynamics learning, inverse kinodynamics learning, behavior cloning, and patch reconstruction with a single representation. \coder~uses a TransformerEncoder to learn the local context of its surroundings by random masking and next patch reconstruction. We show that \coder~achieves better performance across all four different tasks compared to specialized End-to-End models with 77\% fewer parameters. We also show \coder's comparable performance against state-of-the-art kinodynamic modeling and planning approaches in real-world robot deployment. These results underscore the efficacy of \coder~in mitigating overfitting and fostering more robust generalization across diverse environmental contexts and downstream vehicle kinodynamic tasks\footnote{\faGithub~\url{https://github.com/mhnazeri/VertiCoder}}.
\end{abstract}

\section{Introduction} \label{sec::introduction}

Wheeled robots, commonly employed in structured environments such as warehouses, homes, and offices, often encounter limitations when navigating off-road terrain characterized by vertical challenges~\cite{datar2023toward}. In applications like search and rescue and remote inspection, wheeled robots frequently face rocky or uneven terrain filled with obstacles of similar size as the robots. Their inability to negotiate through such vertical protrusions from the ground can disrupt robots' stability, cause wheel slippage, and even lead to damage. 
Such a limitation presents a key obstacle in expanding the operational domain of wheeled robots from controlled settings to unstructured off-road environments.

Recent advancements in wheeled mobility have demonstrated the feasibility of navigating vertically challenging terrain with minimal hardware modifications. Despite the complexity of the terrain, simple all-wheel drive, independent suspensions, and differential locking have proven sufficient in enabling wheeled robots to traverse such environments~\cite{datar2023toward}. Concurrent strides in data-driven robot perception, motion planning, and vehicle control have further contributed to these achievements~\cite{datar2023learning, datar2024terrain, xiao2022motion}.
Among all these successes in traversing vertically challenging terrain~\cite{datar2023learning, datar2024terrain, datar2023toward}, accurate kinodynamic understanding plays a vital role in enabling safe and efficient robot mobility, ranging from forward and inverse kinodynamic modeling~\cite{lee2023learning, xiao2021learning, karnan2022vi, atreya2022high} as well as behavior cloning based on geometric terrain input~\cite{bojarski2016end, pfeiffer2017perception}.

While those end-to-end learning approaches have shown promise, their task-specific nature limits their generalizability to diverse scenarios. In contrast, recent breakthroughs in Self-Supervised Learning (SSL)~\cite{Balestriero2023ACO} suggest the possibility of achieving zero-shot or few-shot learning across various tasks~\cite{o2024open, doshi2024scaling}. For wheeled mobility on vertically challenging terrain, learning a robust and general terrain representation to facilitate kinodynamic understanding will simplify or even improve learning any mobility-related downstream tasks like behavior cloning and 6-DoF kinodynamic modeling. 

\begin{figure}
  \centering
  \includegraphics[width=\columnwidth]{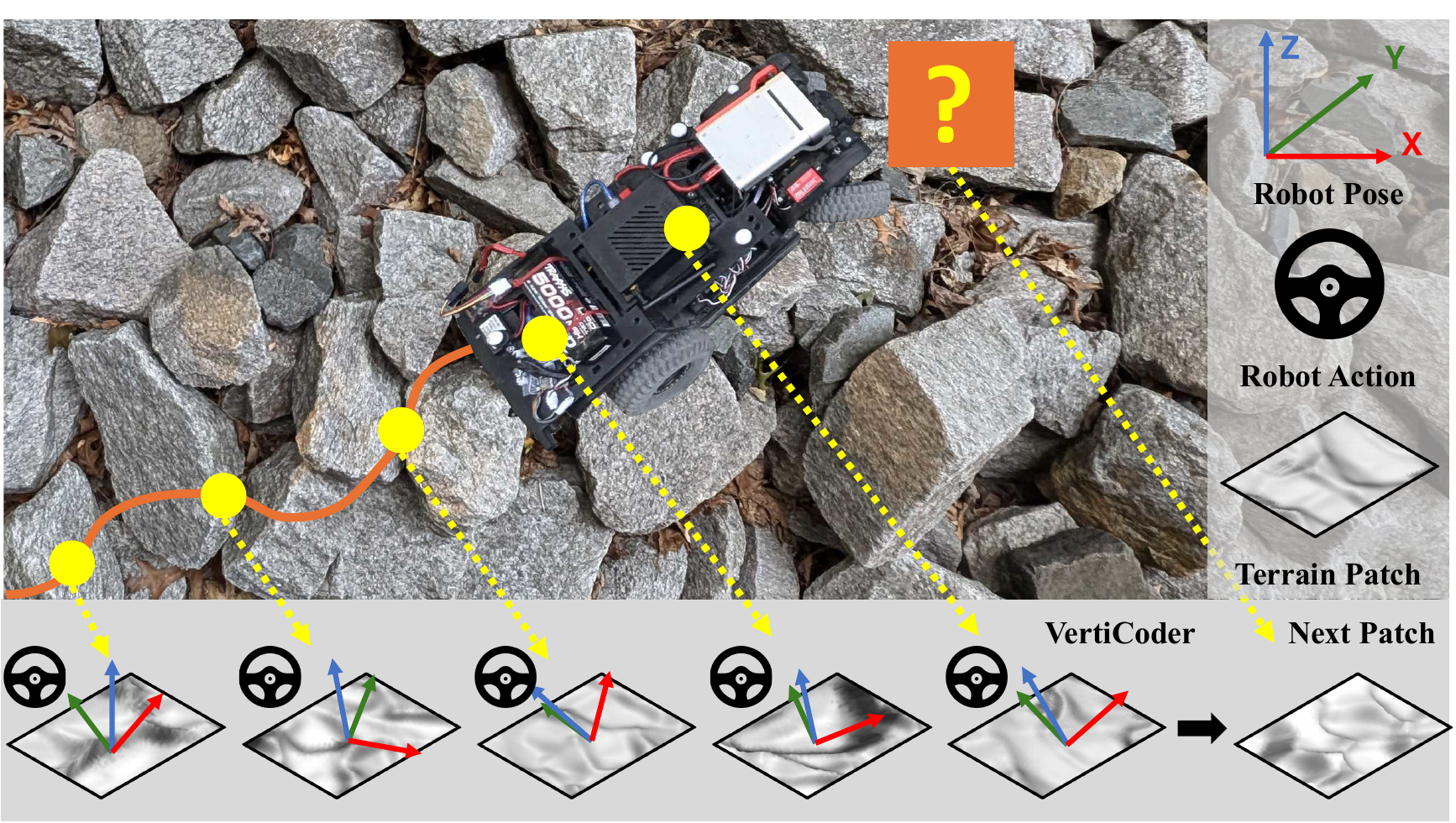}
  \caption{\textbf{\coder}~uses a Transformer to encode a history of terrain patches, robot actions, and robot poses and predicts the representation of the next patch. The learned representation is used in multiple downstream tasks.}
  \label{fig::overview}
\end{figure}
Therefore, in this work, we ask the question: \emph{How can we develop an accurate and generalizable terrain representation for kinodynamic understanding capable of performing multiple tasks with comparable accuracy to its end-to-end counterparts?}
Given the successes of SSL in robotics for solving multiple manipulation tasks~\cite{myers2024policy, zhou2024autonomous}, we introduce \coder, an SSL model that utilizes a transformer encoder~\cite{vaswani2017attention} to reconstruct masked tokens and predict an accurate representation of future terrain patches underneath the robot in vertically challenging terrain.
Inspired by VisionTransformer (ViT)~\cite{dosovitskiy2021an} and VANP~\cite{nazeri2024vanp}, \coder~leverages a TransformerEncoder with an additional context token to reconstruct the masked tokens and next patch underneath the robot. This approach boosts robots' local context awareness. Subsequently, downstream tasks can leverage this context token, which contains historical information of terrain patches, robot poses, and robot actions (Fig.~\ref{fig::overview}), to perform forward kinodynamics learning (FKD), inverse kinodynamics learning (IKD), behavior cloning (BC), and patch reconstruction (PR).

Our experimental results demonstrate better generalization on unseen test environments compared with specialized End-to-End (E2E) models and comparable performance in real-world robot deployment against state-of-the-art models.
The contributions of this work can be summarized as follows:
\begin{itemize}
    \item An SSL model for kinodynamic understanding that can accurately predict future terrain representations;
    \item four different mobility-related downstream tasks from one representation; and 
    \item real-world comparison with previous methods in terms of both learning and robot deployment performance.
\end{itemize}

\section{Related Work}
\label{sec::related_work}

In this section, we first discuss the current approaches in data-driven kinodynamics learning, followed by an exploration of the recent advancements in applications of self-supervised learning in robotics.

\textbf{Learning robot kinodynamics} is a crucial step toward achieving successful navigation in off-road environments~\cite{xiao2022motion}.
While terrain characteristics~\cite{ghosh2018probabilistic} and robot reaction to such terrain can often be predicted~\cite{castro2023does, han2024dynamics} to some extent, the inherent complexity and variability\cite{gibson2023multi} of off-road terrain make it difficult for robots to attain the commanded velocities~\cite{sivaprakasam2021improving}.
To address these challenges, researchers have increasingly focused on learning robot kinodynamics in off-road environments to develop robust control strategies~\cite{pagot2020real} that account for uncertainties and challenges posed by varying terrain conditions~\cite{cai2024evora, cai2024pietra}.
This enables robots to adapt their behavior in real-time during high-speed navigation~\cite{xiao2021learning, karnan2022vi, atreya2022high}, jumping over small hills~\cite{lee2023learning}, mitigating rollovers~\cite{han2023model, talia2024demonstrating}, or even traversing through vertically challenging terrain~\cite{datar2023learning, datar2023toward, datar2024terrain}. 
By learning the robot kinodynamics, researchers have also enabled robots to navigate off-road environments in a risk-aware manner~\cite{cai2022risk,dixit2023step,pokhrel2024cahsor}.
Given the data-intensive nature of many kinodynamics learning methods, recent research has increasingly explored physics-informed~\cite{maheshwari2023piaug, agishev2024monoforce, agishev2024endtoend}, and self-supervised approaches~\cite{pokhrel2024cahsor, datar2024terrain} to reduce the reliance on extensive datasets. \coder~is a self-supervised kinodynamics learning approach that aims at efficiently improving multiple downstream kinodynamic tasks. 

\textbf{Self-Supervised Learning (SSL)} for robotics has emerged as a valuable technique for reducing the reliance on labeled data and improving generalization to unseen environments, a common limitation in supervised learning methods.
SSL techniques leverage intrinsic signals within the data to guide machine learning models, broadly categorized into contrastive~\cite{siva2021enhancing, jung2024vstrong} and information maximization methods~\cite{pokhrel2024cahsor}.
SSL methods have been successfully applied to a variety of off-road robotics tasks, including learning terrain properties from sensor readings such as force sensors~\cite{whereShouldIWalk} and IMU~\cite{sathyamoorthy2022terrapn}, estimating traversability using vision~\cite{FreyRSS23, seo2023learning, jung2024vstrong, zhang2024lift, karnan2024}, pointclouds~\cite{schmid2022self, seo2023scate}, or multi-modal sensor data~\cite{castro2023does}, and acquiring terrain preferences~\cite{kahn2021badgr, sikand2022visual, karnan2024wait}.
Learning terrain representation~\cite{karnan2023sterling, datar2024terrain, pokhrel2024cahsor} is another application of SSL, where the distilled representation is learned based on the property of the terrain which can be then used to perform various downstream tasks in off-road navigation.
\tal~\cite{datar2024terrain}, one of the closest works to \coder,  leverages the entire elevation map and robot pose to reconstruct the terrain patch beneath the robot as a pretext task for learning terrain representations. However, this reliance on a complete map during pre-training can be limiting, as such maps are not always available in real-world scenarios. \coder, in contrast, employs masking and next-patch prediction as its pretext task, thereby eliminating the need for a full map. This approach not only enhances the learned representation's generalizability but also broadens its applicability to a wider range of downstream tasks.

Although first designed for natural language processing, Transformers~\cite{vaswani2017attention} have revolutionized various domains including computer vision~\cite{dosovitskiy2021an, oquab2023dinov2} and robotics~\cite{frey2023fast, jung2024vstrong}.
By leveraging the attention mechanism, Transformers can effectively learn meaningful representations~\cite{Bengio2013, Payandeh2023} from unlabeled data by predicting the next token or masking the surrounding tokens.
Next token prediction has been applied in cross-embodied robot policy learning~\cite{octo_2023, Doshi24-crossformer}. 
The CrossFormer model~\cite{Doshi24-crossformer} used diverse large datasets to learn a wide range of navigation and manipulation tasks for different robot embodiments. However, its context size of 2135 is considerably larger than the more manageable \coder's context size of 61 for a mobile robot with limited onboard computational resources. Moreover, the network's size and complexity make CrossFormer unsuitable for real-time decision-making in vertically challenging environments and deployment on a small Verti-Wheeler platform~\cite{datar2023toward}.

\section{Approach}
\label{sec::approach}
\begin{figure*}
  \centering
  \includegraphics[width=2\columnwidth]{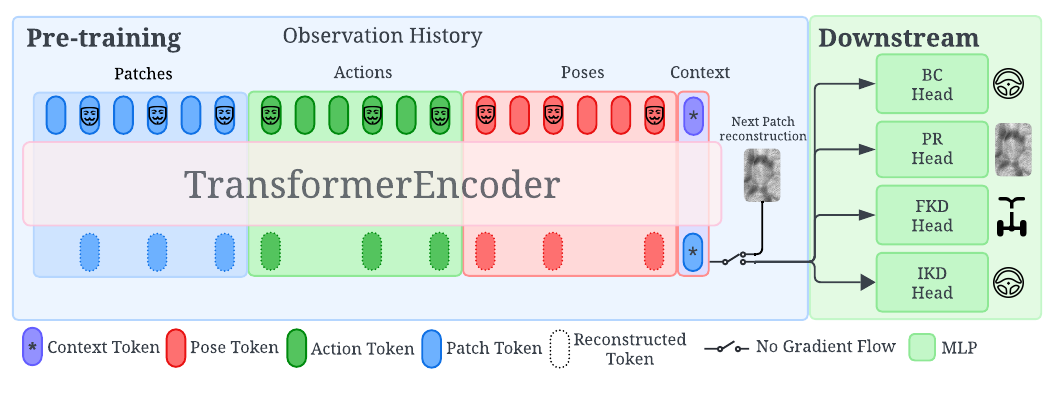}
  \caption{\textbf{\coder~Architecture}. \coder~employs a TransformerEncoder with a context token to predict the representation of the next patch beneath the robot. It achieves this through two simultaneous pretext tasks: predicting the next patch in a sequence of patches, actions, and poses, and reconstructing randomly masked tokens (\mask) within that sequence. Subsequently, we freeze the \coder~and attach multiple downstream task heads to the context token, exclusively training the downstream task heads.}
  \label{fig::coder}
\end{figure*}

Navigating through vertically challenging terrain presents unique challenges in terrain representation learning. In contrast to ego-centric visual navigation, we cannot easily anticipate future obstacles, which complicates decision-making and increases the risk of encountering unforeseen traps. In this section, we first define the pretext task to train the \coder~and how it helps \coder~to anticipate future obstacles. Then we define FKD, IKD, BC, and PR as the downstream tasks (Fig.~\ref{fig::coder}).

\subsection{Pre-training}
Next token prediction has shown great potential in Large Language Models as a pretext objective~\cite{radford2018improving, radford2019language}. Inspired by this approach, we propose leveraging the prediction of the next terrain patch underneath the robot, in conjunction with randomly masked token reconstruction, as a pretext task to align and relate patches with corresponding robot actions and poses. This pretext task facilitates an understanding of the surrounding context and enables the anticipation of future obscured obstacles, thereby enhancing performance and generalization to unseen environments on downstream tasks.

\noindent\textbf{Architecture.} \coder~leverages a TransformerEncoder, inspired by BERT and ViT. It incorporates an additional learnable token, $\mathbf{ctx}$, analogous to BERT's CLS token. For patch encoding, \coder~employs a pre-trained encoder from a SWAE~\cite{kolouri2018sliced}, a component with 3.67 million parameters that constitutes the majority of \coder's model size.

\noindent\textbf{Train.} To train \coder, we use a dataset of $(\mathbf{i}_t, \mathbf{a}_t, \mathbf{p}_t) \in \mathcal{O}$, where $\mathbf{i}_t$ is a terrain patch of size $40\times40$ beneath the robot at time $t$, $\mathbf{a}_t$ is the action (linear ($v_t$) and angular ($\omega_t$) velocities) taken by the robot on the patch $\mathbf{i}_t$, $\mathbf{p}_t$ is the robot pose in $\mathbb{SE}(3)$ on the patch $\mathbf{i}_t$, and $\mathcal{O}$ is the observation space. We pass a history of 20 timesteps as an input to \coder. We use three separate encoders to map $(\mathbf{i}_t, \mathbf{a}_t, \mathbf{p}_t)$ to consequent tokens. For terrain patches, $\mathbf{i}_t$, we use the frozen SWAE encoder, denoted by $\text{SW}_{enc}$, to encode the patches into patch tokens $\tau_{t}^{\mathbf{i}}$. For actions $\mathbf{a}_t$, we employ a linear model, $\text{enc}_{\mathbf{a}}$, to map actions into action tokens, denoted as $\tau_{t}^{\mathbf{a}}$. And finally, to tokenize robot poses $\mathbf{p}_t$, we use another linear model, $\text{enc}_{\mathbf{p}}$, to map poses into pose tokens, denoted by $\tau_{t}^{\mathbf{p}}$. Then, we prefix the observation tokens $(\tau^{\mathbf{i}}_i, \tau^{\mathbf{a}}_i, \tau^{\mathbf{p}}_i)_{i=t-19}^t$ with a context token $\mathbf{ctx}$ and randomly masked 75\% of the tokens before passing them to \coder. The training of the \coder~is then accomplished by calculating and minimizing the mean squared error between the predicted and actual terrain patch embeddings.

\subsection{Downstream Tasks}
\textbf{Forward Kinodynamics Learning (FKD):} We follow the definition of Datar \emph{et al.}~\cite{datar2024terrain} for FKD and adopt a discrete vehicle forward kinodynamic model:

\begin{equation}
    \mathbf{p}_{t+1} = f_\psi(\mathbf{ctx}), 
    \nonumber
\end{equation}
where $\mathbf{ctx}$ is \coder's context token, $\mathbf{p}_{t+1}$ is the next robot pose, and $\psi$ is learnable parameters. 
We aim to show that the learned representation possesses all the necessary information to predict the subsequent pose without explicitly taking in any other information. 
However, for the end-to-end model used for comparison, we enhance the input by explicitly incorporating the current action $\mathbf{a}_{t}$ and pose $\mathbf{p}_{t}$ in addition to the patch embedding.

The FKD model can be used in sampling-based motion planners to produce potential future vehicle trajectories, which will be evaluated based on a cost function, to move the robot to its goal while minimizing the chance of failure on vertically challenging terrain (e.g., rollover or getting stuck). 

\textbf{Inverse Kinodynamics Learning (IKD):}
We follow the definition of Karnan \emph{et al.}~\cite{karnan2022vi} for IKD and use the context token as input to the vehicle inverse kinodynamic model:
\begin{equation}
    \mathbf{a}_{t} = f_\epsilon(\mathbf{ctx}, \mathbf{p}_{t+1}), 
    \nonumber
\end{equation}
where $\mathbf{a}_{t}$ is the current action to be taken by the robot in order to reach $\mathbf{p}_{t+1}$,  the robot's desired next pose including translations and rotations along the $x$, $y$, and $z$ axes. $\epsilon$ is learnable parameters. 

The IKD model can be used with a global planner that constantly produces the robot's desired next state to move the robot toward its goal safely. 

\textbf{Behavior Cloning (BC):}
We follow the definition of Nazeri \emph{et al.}~\cite{nazeri, nazeri2024vanp} for Behavior Cloning: 
\begin{equation}
\mathbf{a}_{t} = \pi_{\zeta}(\mathbf{ctx}, g), 
\nonumber
\end{equation}
where $\pi$ is a controller policy parameterized by $\zeta$ and $g$ is a goal. If we do not include the goal, the policy learns to explore or drive forward without a goal, depending on the training data.

\textbf{Patch Reconstruction (PR):}
To reconstruct the next patch, the SWAE decoder reconstructs the subsequent patch from the context token. This process is formulated as:
\begin{equation}
    \mathbf{i}_{t+1} = \text{SW}_{dec}(\mathbf{ctx}), \nonumber 
\end{equation}
where $\mathbf{i}_{t+1}$ is the next terrain patch and $\text{SW}_{dec}$ denotes the decoder (in contrast to the SWAE encoder $\text{SW}_{enc}$ used in \coder~pre-training). Notice that the PR downstream task is also the same \coder~pre-training task.

All downstream tasks are learned by minimizing supervised losses between the prediction and the ground truth of next pose $\mathbf{p}_{t+1}$, current action $\mathbf{a}_{t}$, and next patch $\mathbf{i}_{t+1}$ for FKD, IKD, BC, and PR, respectively. For the PR task, we also use the peak signal-to-noise ratio (PSNR) metric to evaluate the quality of the reconstructed images. 

\subsection{Implementations}\label{sec:implementations}
In terms of hardware configuration, this work employs an open-source Verti-4-Wheeler (V4W) platform, as described by Datar \emph{et al.}~\cite{datar2023toward}. The V4W is equipped with a Microsoft Azure Kinect RGB-D camera and an NVIDIA Jetson Xavier processor. Additionally, it features low-gear drive and lockable front and rear differentials, which significantly enhance its mobility on vertically challenging terrain. On the software side, \coder~is implemented with PyTorch and trained on a single A5000 GPU with 24GB memory.

\textbf{Architecture:} \coder~consists of a TransformerEncoder with an additional context vector like BERT~\cite{Devlin2019}, ViT~\cite{dosovitskiy2021an}, and VANP~\cite{nazeri2024vanp} with four layers and four heads to produce context embedding $\mathbf{ctx} \in \mathbb{R}^{128}$. The context size of the Transformer is 61 including 20 patches, 20 actions, 20 poses, and 1 context token. Using the Transformer along with masking and next patch prediction allows the model to understand the local context of the terrain and how a variation in terrain height and robot action can affect the robot pose over time. For downstream tasks, we use separate small multi-layer perceptrons with $[128, 64, 32, 16, n]$ layers, where $n$ is the downstream output dimension to map the context token to the target task space.

\textbf{Optimization:} For both pretext and downstream training we use the AdamW optimizer~\cite{Loshchilov2019}. We pre-train \coder~for 300 epochs with a batch size of 512. We then train the downstream heads for 50 epochs for both \coder~ and its end-to-end counterpart with a batch size of 32 for a fair comparison. We observe that using a larger batch size than 32 for the downstream heads causes the model to converge to the mean of the data.

\textbf{Freezing and Fine-tuning:} We employ two versions of the training process for downstream tasks. First, we freeze the weights of the \coder~and train only the downstream task-specific heads. This strategy allows us to evaluate the expressivity and generalizability of \coder's learned common representations. Second, in the fine-tuned version, we unfreeze the \coder~weights, allowing them to be updated in conjunction with the downstream heads using the downstream loss. This fine-tuning process is implemented to enhance performance, as it enables the \coder~model to adapt its pre-learned common representations to the specific nuances of each downstream task, potentially leading to more task-specific feature extraction and improved overall performance.

\textbf{End-to-End Architecture:}
For the specialized E2E models, we use Resnet-18~\cite{He2015} as the patch encoder and attach multi-layer perceptrons with $[512, 256, 512, 64, n]$ layers as the task heads, where $n$ is the dimension of downstream output. A notable distinction between these end-to-end models and the \coder~lies in the input modality where the former explicitly incorporates additional information, such as current pose and action to predict next pose for FKD and to predict next patch for PR, whereas the latter exclusively utilizes the context token $\mathbf{ctx}$, which implicitly includes such information through pre-training, as input to the downstream heads. 
A comparison of \coder~against such end-to-end architectures with explicit task-specific inputs aims to showcase the effectiveness of \coder~in extracting relevant and common features to implicitly use them for different downstream tasks during the same pre-training process. 

\begin{figure}
  \centering
  \includegraphics[width=0.75\columnwidth]{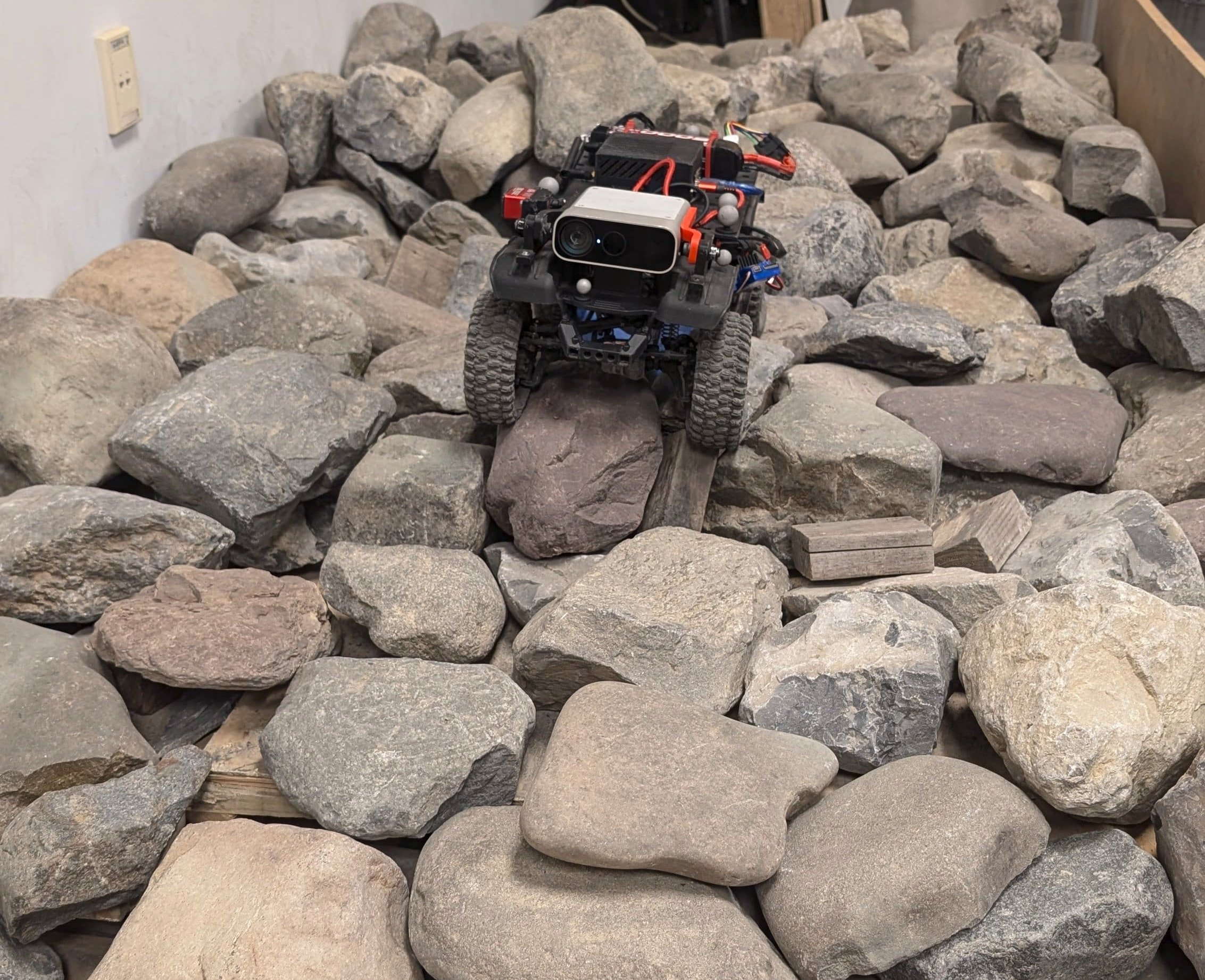}
  \caption{\textbf{Rock Testbed and V4W used for Data Collection and Experiments}. The modularity of the testbed allows diverse rock configurations for training and evaluation.}
  \label{fig::testbed}
  \vspace{-0.5em}
\end{figure}

\textbf{Dataset:} We use the dataset employed by \tal~\cite{datar2024terrain}, which is collected on a rock testbed measuring 3.1 m $\times$ 1.3 m with a maximum height of 0.6 m (Fig.~\ref{fig::testbed}). The rock testbed's modular design allows for easy reconfiguration, facilitating mobility experiments across various terrain arrangements. Given that ground vehicle dynamics are predominantly influenced by terrain topology, and considering the substantial computational requirements associated with full 3D mapping, a 2.5D terrain elevation map is employed to construct the patches beneath the vehicle~\cite{mikielevation2022}. The dataset includes 30 minutes of teleoperating the robot on the rock testbed and an additional 30 minutes on a planar surface. The dataset is divided using a 9:1 ratio for training and testing purposes. The dataset encompasses visual inertial odometry for vehicle state estimation, elevation maps generated from depth images, and teleoperated vehicle control inputs, including throttle and steering commands. The rock testbed dataset captures a diverse range of 6-DoF vehicle states, including instances of vehicle rollover and immobilization.

\section{Experiments}
\label{sec::experiments}
\begin{table*}[t]
    \centering
    \caption{\textbf{Downstream Performance.} Comparison of the \coder~performance on four different tasks compared with end-to-end models on seen/unseen data.~\faSnowflake~denotes frozen~\coder~backbone, and~\faFire* denotes fine-tuned model.}
    \begin{NiceTabular}{lccccccccc}
    \CodeBefore
    \rowcolor{gray!15}{7}
    \Body
    \toprule
    \textbf{Method}   & \Block[c]{1-2}{\textbf{\#Params (M)~$\downarrow$}} & & \Block[c]{1-2}{\textbf{FKD~$\downarrow$}} &  & \Block[c]{1-2}{\textbf{IKD}~$\downarrow$} &  & \Block[c]{1-2}{\textbf{BC}~$\downarrow$}& & \textbf{PR}~$\uparrow$ \\
                                         \cmidrule(rl){2-3}    \cmidrule(rl){4-5}  \cmidrule(rl){6-7} \cmidrule(rl){8-9} \cmidrule(rl){10-10}
                                         & Learnable & Total & Train & Test & Train & Test & Train & Test    & Test \\
    \midrule                      
    \textbf{\textsc{EndToEnd-18}} & $\approx11.47$& $\approx11.47$       & \textbf{0.002} & 0.031   & \textbf{0.012} & 0.048    & \textbf{0.015} & 0.259       & 18.91   \\

    \midrule
    \textbf{\textsc{EndToEnd-50}} & $\approx 26.42$& $\approx26.42$           & 0.002 & 0.031   & 0.039 & 0.117    & 0.017 & 0.292       & -   \\
    \midrule
    \textbf{\textsc{EndToEnd-EfficientB4}} & $\approx21.04$& $\approx21.04$           & 0.010 & 0.032   & 0.037 & 0.120    & 0.108 & 0.283       & -   \\
    \midrule
    \textbf{\coder~\faSnowflake} & \textbf{0.01} & \textbf{2.71} & 0.007 & \textbf{0.009}  & 0.073 & \textbf{0.045}    & 0.172 & \textbf{0.159}       & \textbf{25.65}   \\
    \midrule 
    \textbf{\coder~\faFire*} & 0.88 & 2.71 & 0.003 & 0.002  & 0.008 & 0.002    & 0.032 & 0.017       & 28.093   \\

    \bottomrule
    \end{NiceTabular}
    
    \label{tab:result}
\end{table*}
 
To validate \coder's efficacy on the four specified downstream tasks while relying solely on the context token, we conduct a comparative analysis between \coder's predictions and those of specialized E2E models. Furthermore, we demonstrate the practical applicability of \coder~by deploying it on the V4W robot to navigate through vertically challenging terrain using BC. This dual approach of comparative evaluation and practical implementation serves to comprehensively assess the versatility and performance of the \coder~framework across diverse task domains.

\subsection{\coder~Experiments}
\textbf{Accuracy across Different Downstream Tasks:}
A comparative analysis of \coder's accuracy across four different downstream tasks, FKD, IKD, BC, and PR, against specialized E2E models reveals noteworthy findings, as illustrated in Table~\ref{tab:result}. E2E-18, E2E-50, and E2E-EfficientB4 leverage Resnet-18, Resnet-50, and EfficientNet-B4 as their backbone respectively. The E2E models demonstrate a tendency to overfit the training data, resulting in suboptimal performance when confronted with unseen data. Increased depth in E2E-50 and E2E-EfficientB4 does not help the model to understand the complexities and cause the model to overfit more than the shallower E2E-18 model In contrast, both the frozen and fine-tuned versions of \coder~exhibit superior generalization capabilities across tasks and unseen environments. The enhanced generalization of \coder~can be attributed to its use of masking and next token prediction as pretext tasks during the pre-training phase. Note that \coder~downstream heads only receive the context token as input while E2E models have access to additional information as mentioned in Sec.~\ref{sec::approach}. Since PR is also part of \coder's pretext task and its downstream head, i.e., $\text{SW}_{dec}$, does not need to be re-trained, it does not have a column for training loss on seen data. 

\begin{table}
    \centering
    \caption{\textbf{Real-World Robot Deployment.} Comparison between \coder~BC and E2E BC.}
    \label{tab::bc_exp}
    \begin{NiceTabular}[columns-width = 0.25cm,rules/width=1pt]{lcc}
    \toprule
    & \coder~BC & E2E BC\\
    \midrule
    Success Rate $\uparrow$&  \textbf{8/10} & 7/10\\
    Average Time $\downarrow$&14.47$\pm$1.81 &  \textbf{12.28$\pm$2.69} \\

\bottomrule
\end{NiceTabular}
\end{table}

Furthermore, \coder's superior resistance to overfitting is achieved using a mere 23\% of the parameters employed by the E2E model. Notice that the difference between Learnable (0.88M) and Total (2.71M) parameters for the fine-tuned \coder~is due to the always frozen $\text{SW}_{enc}$ parameters derived from SWAE encoder and decoder pre-training prior to \coder~pre-training. This also shows that most of the \coder's parameters are coming from $\text{SW}_{enc}$. This marked reduction in parameter count, coupled with improved generalization, underscores \coder's efficient training and inference time. 
Moreover, this approach yields an additional benefit in the form of reduced training time for downstream tasks. The compact yet expressive representations learned by \coder~outperform or produce comparable results against four different specialized E2E models and facilitate more efficient fine-tuning, as the model requires less adaptation to new task-specific objectives.

\begin{table}[t]
    \centering
    \caption{\textbf{Mask Percentage Ablation Study.} Comparison of different percentages of masking on FKD and PR.}
    \begin{NiceTabular}{lccc}
    \toprule
    \diagbox{\scriptsize{\textbf{Mask}}}{\scriptsize{\textbf{Task}}}   & \Block[c]{1-2}{\textbf{FKD~$\downarrow$}} & & \textbf{PR~$\uparrow$}\\
                                         \cmidrule(rl){2-3}               \cmidrule(rl){4-4}
                                         & Train & Test &                 Test  \\
    \midrule                              
    \textbf{\textsc{75\%}} &          0.058 & \textbf{0.050}                    & \textbf{22.775}   \\

    \midrule
    \textbf{\textsc{90\%}} &          \textbf{0.006} & 0.055                    & 16.623    \\

    \bottomrule
    \end{NiceTabular}
    
    \label{tab:masking}
    \vspace{-0.5em}
\end{table}

\textbf{On-robot deployment:}
We deploy the \coder~with BC on the V4W. We conduct a comparative analysis of \coder's performance against E2E BC. We utilize the reported results from \tal~as our baseline for comparison and like \tal~\cite{datar2024terrain}, we set the goal across the rock test bed. As demonstrated in Table~\ref{tab::bc_exp}, \coder~BC outperforms E2E BC in terms of a higher success rate with a slower, but more steady speed. However, we observe that \coder~BC exhibits a performance gap when compared to \wmvct~\cite{datar2023learning} and \tal~\cite{datar2024terrain}, two specialized kinodynamic modeling approaches used in conjunction with sophisticated sampling-based motion planners. This suggests that while \coder~offers promising results in the real world, there remains room for further enhancement to achieve parity with state-of-the-art methods like \wmvct~and \tal. Future research could explore the integration of complementary techniques to further refine the representation and narrow this performance gap.

\subsection{Ablation Studies}
To investigate the most effective approach to train \coder~we experiment with a series of ablation studies. We choose the FKD accuracy and PR PSNR accuracy for ablations. We collect a new 30-minute dataset on the rock testbed which is split into 9:1 ratio for the train and test data. The ablation dataset is collected using a motion capture system for odometry while keeping the rest of the inputs the same. 

\begin{table}[t]
    \centering

    \caption{\textbf{Token Ordering Ablation Study.} Comparison of different ways to feed tokens to \coder~on FKD and PR.}
    \begin{NiceTabular}{lccc}
    \toprule
    \diagbox{\scriptsize{\textbf{Ordering}}}{\scriptsize{\textbf{Task}}}   & \Block[c]{1-2}{\textbf{FKD~$\downarrow$}} & & \textbf{PR~$\uparrow$}\\
                                         \cmidrule(rl){2-3}               \cmidrule(rl){4-4}
                                         & Train & Test &                 Test  \\
    \midrule                              
    \textbf{Interleaving} &          0.065 & 0.052                    & 22.306   \\ 

    \midrule
    \textbf{Sequential} &          \textbf{0.058} & \textbf{0.050}                    & \textbf{22.775}    \\ 
    \bottomrule
    \end{NiceTabular}
    
    \label{tab:order}

\end{table}

\textbf{Role of Masking:} To investigate the impact of masking on accuracy, we conduct experiments with varying masking percentages. As detailed in Table~\ref{tab:masking}, we evaluate the model's performance with masking percentages of 75\% and 90\%, in line with recommendations from previous studies~\cite{Devlin2019, assran2023self}. Our findings reveal that masking 90\% of the tokens hinders the model's ability to effectively grasp the context while masking 75\% of the tokens facilitates a more effective representation of the surrounding information.

\textbf{Order of tokens:} Furthermore, we explore the influence of token ordering within the representation. We experiment with both interleaved token arrangements, $\{\tau_{t-19}^{\mathbf{i}}, \tau_{t-19}^{\mathbf{a}}, \tau_{t-19}^{\mathbf{p}}, \ldots, \tau_{t}^{\mathbf{i}}, \tau_{t}^{\mathbf{a}}, \tau_{t}^{\mathbf{p}}\}$, and sequential ordering, $\{\tau_{t-19}^{\mathbf{i}}, \ldots, \tau_{t}^{\mathbf{i}}, \tau_{t-19}^{\mathbf{a}}, \ldots, \tau_{t}^{\mathbf{a}}, \tau_{t-19}^{\mathbf{p}}, \ldots, \tau_{t}^{\mathbf{p}}\}$. The results, presented in Table~\ref{tab:order}, demonstrate superior model performance in both FKD and PR tasks when employing sequential ordering. This suggests that a sequential arrangement of tokens contributes to enhanced alignment between the different modalities.

\subsection{Discussions}
The disparity in performance of E2E models and \coder~can be attributed to the distinct architectural and training paradigms employed by each model. The E2E models, while potentially excelling in capturing intricate patterns within the training set, appear to lack the robustness required for effective generalization. Conversely, \coder's architecture coupled with its training methodology---involving both a frozen and unfrozen representation---seems to confer enhanced adaptability to unfamiliar scenarios.
The frozen version of \coder, leveraging pre-trained representations without task-specific adjustments, demonstrates the inherent transferability of its learned features. The fine-tuned variant further refines these representations, striking a balance between retaining generalizable knowledge and adapting to task-specific nuances. The \coder's self-supervised learning technique appears to be instrumental in cultivating a more robust and generalizable kinodynamic representation of the underlying data distribution.

\subsection{Limitations}
In our analysis, we identify several key limitations of the \coder. Although \coder~demonstrates superior performance on the test dataset compared to E2E models, it falls short of surpassing state-of-the-art models, \wmvct~and \tal, in real-world scenarios. This indicates that while \coder's representation helps prevent overfitting on the training data, it still lacks the expressivity required to generalize effectively to real-world situations.
We attribute this limitation to \coder's reliance on local context awareness (i.e., its dependence on the patch directly beneath the robot), which excludes crucial global information from the map. This restricted access to global context hampers \coder's ability for long-horizon planning. Furthermore, when the robot moves at a slow pace, there is minimal change in information between consecutive terrain patches. This lack of temporal variation can hinder the model's learning process.
While data augmentation can potentially alleviate this issue, crude augmentations on elevation maps risk significantly altering the context, leading to inaccurate predictions in downstream tasks. Therefore, it is essential to develop more sophisticated augmentation techniques that preserve contextual integrity while providing sufficient variation for effective learning.

\section{Conclusions}
In this paper, we introduce \coder, a self-supervised model designed to learn a general kinodynamic representation via a TransformerEncoder that can be applied across four distinct tasks for robot mobility on vertically challenging terrain: forward kinodynamics learning, inverse kinodynamics learning, behavior cloning, and patch reconstruction. This versatility is achieved through a pretext task involving random masking and next-patch reconstruction, promoting an effective representation of the robot's local context. Our experiments demonstrate that \coder~not only outperforms specialized end-to-end models across all four tasks while utilizing 77\% fewer parameters, but also exhibits comparable performance to state-of-the-art approaches in real-world deployments. These results highlight the effectiveness of \coder's self-supervised approach in mitigating overfitting and facilitating robust generalization across diverse and challenging tasks and environments.
\label{sec::conclusions}

\section*{Acknowledgments}
This work has taken place in the RobotiXX Laboratory at George Mason University. RobotiXX research is supported by National Science Foundation (NSF, 2350352), Army Research Laboratory (ARL, W911NF2220242, W911NF2320004, W911NF2420027, W911NF2520011), Air Force Research Laboratory (AFRL) and US Air Forces Central (AFCENT, GS00Q14OADU309), Google DeepMind (GDM), Clearpath Robotics, and Raytheon Technologies (RTX).

\bibliographystyle{IEEEtran}
\bibliography{IEEEabrv,references}
\end{document}